\documentclass[11pt]{article}

\usepackage{acl}

\usepackage{times}
\usepackage{latexsym}
\usepackage[T1]{fontenc}
\usepackage[utf8]{inputenc}
\usepackage{microtype}
\usepackage{inconsolata}
\usepackage{graphicx}
\usepackage{booktabs}

\usepackage{amsmath}
\usepackage{amssymb}
\usepackage{multirow}
\usepackage{array}

\title{From Noisy Retrieval to Decision Evidence: Silver Evidence Demonstrations for In-Context Learning}

\author{
\textbf{Masoud Monajatipoor} \thanks{Equal contribution.}
\quad
\textbf{Fazlolah Mohaghegh} \footnotemark[1]
\quad
\textbf{Hamid Reza Hassanzadeh} \\
\textbf{Akash Uttamrao Bhoite} \quad
\textbf{Robert E. Tillman} \\[6pt]
Optum AI \\[3pt]
\texttt{\{masoud\_monajatipoor,ehsan.mohaghegh,}\\
\texttt{hamid.hassanzadeh,akash.bhoite,robert.tillman\}@optum.com}
}

\begin{document}

\maketitle

\begin{abstract}

RAG and retrieval-augmented few-shot ICL can struggle when decision-relevant evidence is buried in long, noisy context. Retrieved passages and demonstrations may contain redundant or distracting information. Although expert-labeled development examples may identify the text supporting each decision, these annotations are typically used only as answer-side rationales rather than as supervision for filtering noisy retrieval. We propose \emph{silver-evidence demonstrations}, which replace raw retrieved passages in ICL examples with generated, target-conditioned evidence summaries. For each labeled development example, an LLM generates silver evidence without access to the expert label or gold evidence; this evidence is then paired with the expert label and gold evidence where available. At test time, a single structured call generates silver evidence from retrieved passages and then predicts the final label and decision evidence. Across EI 2.0, SciFact, synthetic policy retrieval, and HICRIC/IMR data, our full method improves over equally sized gold-only ICL (label-only for policy) on three of four datasets and improves evidence F1 on all annotated datasets. It also achieves higher mean label accuracy than InstructRAG-style ICL on all four datasets. These results suggest that replacing demonstration-side retrieved passages with compact, target-conditioned silver evidence is particularly useful for decisions based on long and noisy text, where it reduces distracting context while preserving decision-relevant signals and expert supervision.

\end{abstract}

\begin{figure*}[!t]
\centering
\includegraphics[width=\textwidth]{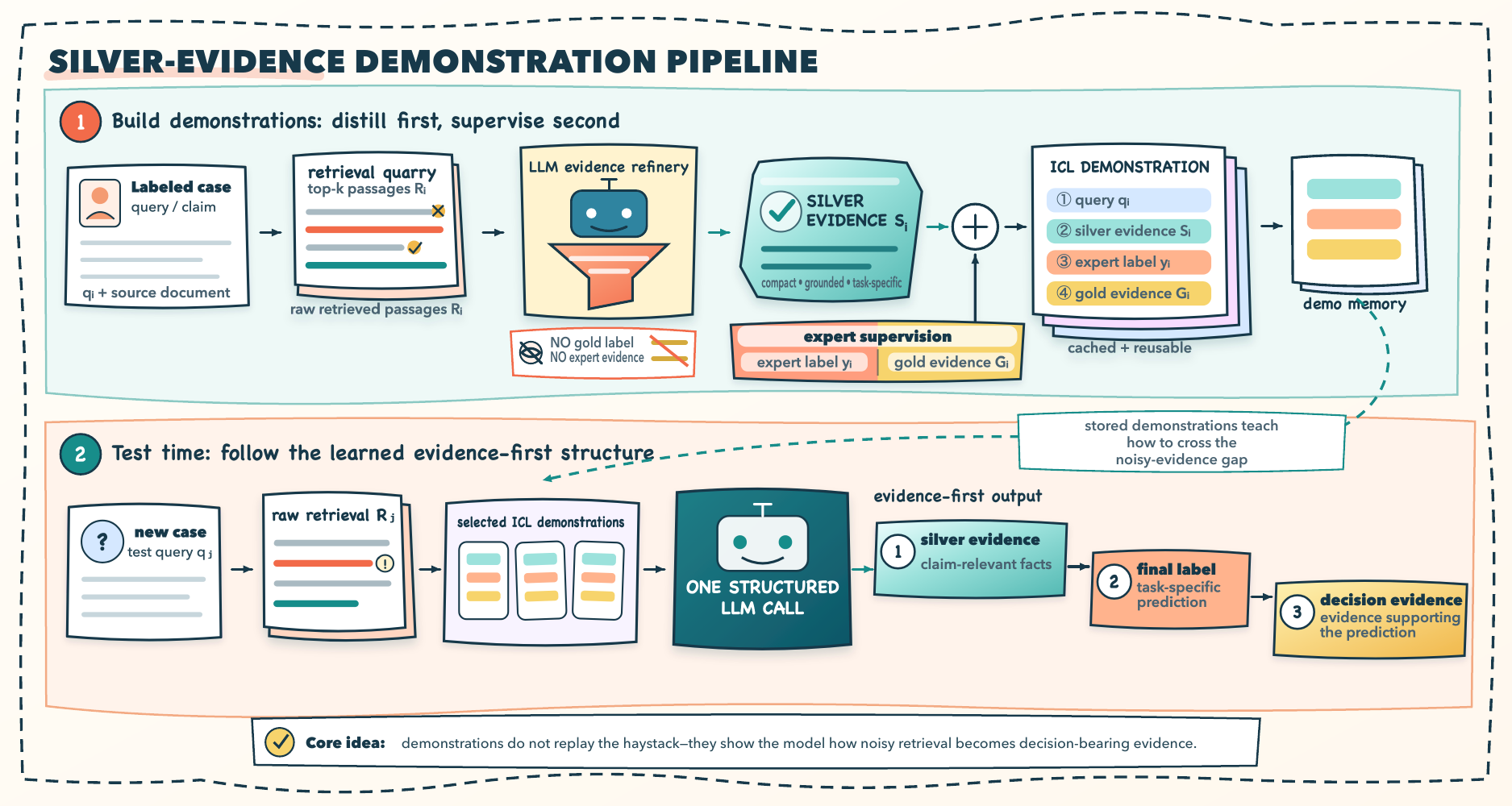}
\caption{End-to-end silver-evidence pipeline. Stage 1 constructs demonstrations by generating silver evidence from retrieved passages without access to expert labels or expert gold evidence. Stage 2 uses those demonstrations to prompt the model to produce silver evidence before predicting the final label and evidence.}
\label{fig:silver-pipeline}
\end{figure*}

\section{Introduction}

Many high-stakes language understanding tasks require both a decision and a justification grounded in source text. This is especially important in clinical and medical settings, where evidence synthesis, scientific claim verification, and coverage or prior-authorization workflows connect a decision to trial sentences, scientific rationales, or policy language. Useful evidence is often a small subset of a larger retrieved context, and it is part of the decision product rather than merely an explanation artifact. Users and reviewers may need to know which source text drove a support, rejection, approval, denial, or pending decision for audit, appeal, expert review, and error correction.

RAG addresses this problem by retrieving candidate passages for generation \cite{lewis2020rag}, while few-shot ICL supplies labeled demonstrations \cite{brown2020language}. Their combination remains vulnerable when demonstrations and test inputs are long and noisy: retrieved passages can be redundant, irrelevant, or distracting, and prompt limits restrict how many examples can be shown. Although development examples may identify the text supporting a decision, conventional ICL presents this annotation only as an answer-side rationale, giving little guidance on transforming broad retrieval into expert-style evidence.

This paper studies a two-stage alternative, summarized in Figure~\ref{fig:silver-pipeline}. We first transform each development example's retrieved passages into \emph{silver evidence}: a concise, decision-focused selection and synthesis of source-grounded information, conditioned on the target claim, intervention, policy criterion, or decision question but not its label or expert evidence.
Demonstrations replace full retrieval with this decision-focused evidence and pair it with the label and, when available, gold evidence. At inference, one LLM call transforms the target's raw retrieval into silver evidence before predicting the label and gold-style evidence. This creates an intentional asymmetric ICL setting: demonstrations show compact evidence paired with expert supervision, whereas target examples retain raw retrieval and require the model to construct the corresponding evidence representation before making a decision.

Our contributions are threefold: an asymmetric demonstration format that replaces raw demonstration retrieval with label-blind, task-conditioned silver evidence paired with expert supervision; a single-call inference workflow that generates silver evidence from target retrieval before predicting the label and gold-style evidence; and comparisons with retrieval-only, gold-only, silver-only, oracle, and InstructRAG-style demonstrations across LLM backends. We evaluate Evidence Inference 2.0 \cite{deyoung2020evidence}, SciFact \cite{wadden2020scifact}, synthetic prior-authorization policy retrieval \cite{sharifullin2026offline}, and an adapted HICRIC/IMR appeal-policy setting \cite{gartner2024hicric}. 
We first compare backends using the full method. GPT-5.5 achieves the strongest aggregate label performance across the four datasets and is therefore used for the controlled system comparisons. Compared with gold-only ICL, its gains are largest on policy retrieval and SciFact, smaller on HICRIC/IMR, and minimal on Evidence Inference. The full method achieves higher accuracy than InstructRAG on all four datasets with slightly lower policy Macro-F1.

\section{Related Work}

\paragraph{Retrieval-augmented generation.}
RAG combines parametric language models with retrieved non-parametric memory for knowledge-intensive generation \cite{lewis2020rag}. Prior work studies retrieval during pretraining, dense and late-interaction passage retrieval, memory augmentation, document aggregation, in-context retrieval, and adaptive retrieval with reasoning or self-critique \cite{guu2020realm,karpukhin2020dpr,khattab2020colbert,khandelwal2020knnlm,borgeaud2022retro,izacard2021fid,ram2023incontext,shi2024replug,yao2022react,asai2023selfrag,jiang2023flare}. Yet more context is not always better: models may underuse long contexts or require selective retrieval \cite{liu2024lost,mallen2023trust}. RECOMP, Chain-of-Note, and SEER address compression or evidence extraction \cite{xu2024recomp,yu2024chain,zhao2024seer}; our contribution is a label-blind, asymmetric ICL format that pairs generated evidence with expert supervision without training a compressor.

\paragraph{In-context learning and intermediate reasoning.}
ICL behavior depends on example selection, order, semantic similarity, format, calibration, and prompt budget \cite{xie2022bayesian,dong2024survey,min2022rethinking,liu2022makes,lu2022fantastically,rubin2022learning,zhao2021calibrate,schick2021pet,gao2021lmbff,shin2020autoprompt}. Long retrieved passages leave room for fewer demonstrations, motivating compact intermediate representations \cite{liu2024lost,ram2023incontext}. Chain-of-thought methods show that intermediate text can aid reasoning, but its content and structure matter \cite{wei2022cot,kojima2022zeroshot,zhou2023least,wang2023selfconsistency,wang2022cot}. Closest to our work, InstructRAG uses answer-conditioned rationales while retaining retrieved documents in ICL demonstrations \cite{wei2025instructrag}; our label-blind silver evidence however replaces demonstration retrieval and is paired with expert evidence when available.

\paragraph{Evidence and rationale supervision.}
Rationale extraction supports interpretability \cite{lei2016rationalizing}, while e-SNLI and ERASER connect predictions to explanations or annotated evidence \cite{camburu2018esnli,deyoung2019eraser}. Prior work warns that plausible explanations are not necessarily grounded in the source text or faithful to the model's decision process \cite{jain2019attention,wiegreffe2019attention,jacovi2020faithfulness}. Evidence Inference and biomedical resources support evidence-centered clinical processing \cite{deyoung2020evidence,nye2018corpus,romanov2018mednli,jin2019pubmedqa,lee2020biobert,beltagy2019scibert}; SciFact, FEVER, and related datasets extend this to scientific and public-health verification \cite{wadden2020scifact,thorne2018fever,hanselowski2018ukp,aly2021feverous,atanasova2020factexplanations,kotonya2020publichealth}. Policy, legal, and contract datasets pose related long-document decision problems \cite{sharifullin2026offline,gartner2024hicric,chalkidis2020legalbert,hendrycks2021cuad,koreeda2021contractnli}. These settings motivate our distinction between retrieval-derived silver evidence and expert gold evidence.

\section{Methodology}

\subsection{Problem Setup}

Each example consists of a target query $q$, retrieved passages $R=\{r_1,\ldots,r_k\}$, a label $y$, and optionally expert evidence $G=\{g_1,\ldots,g_m\}$. Dataset-specific retrieval is held fixed across systems; our method begins from $(q,R)$ rather than optimizing retrieval. Queries are clinical prompts, scientific claims, policy requests, or appeal backgrounds, with dataset-specific label spaces. We predict $\hat y$ and, where available, supporting evidence $\hat G$.

\subsection{Silver Evidence Generation}

For each labeled training or validation example, we call an LLM with only the query and retrieved passages:
\[
S_i = f_{\theta}^{silver}(q_i, R_i).
\]
The prompt is target-conditioned: depending on the dataset, it identifies the decision question, claim, policy criterion, or appeal background and then presents the retrieved passages. Available titles, headers, or sentence identifiers, 
are retained. The prompt excludes the gold label and evidence and requests concise, source-supported snippets organized around the decision, making this call retrieval summarization rather than label prediction.

Silver evidence is generated once for the selected example split and cached for all target prompts. Reusing this cache prevents test-time examples from changing demonstration evidence and makes construction reproducible.

The evidence representation is intentionally structured rather than a free-form summary. When retrieval provides a document title, policy heading, category, sentence identifier, or other source metadata, the silver item preserves that key together with the supporting text. This allows the final prompt to learn two related operations: filtering the retrieved material and expressing the result in the evidence format expected by the downstream evaluator. The generator is instructed to avoid broad background statements, unsupported inferences, and label-specific wording. It may select multiple snippets when a decision depends on complementary findings, but each item must remain traceable to the supplied retrieval.

\subsection{Demonstration Construction and Leakage Control}

Each ICL demonstration is built after silver evidence is cached:
\[
(q_i, S_i, y_i, G_i),
\]
where $G_i$ is omitted when unavailable. Unlike retrieval-augmented ICL, demonstrations expose the compact evidence representation rather than raw passages, reducing prompt cost and aligning generated evidence with expert supervision. Because $R_i$ is replaced rather than retained, the pair $(R_i,S_i)$ is not shown directly.

The construction enforces leakage controls: silver generation excludes $y_i$ and $G_i$; demonstrations come only from the prepared training split; and target labels/evidence are evaluation-only. 
Oracle evidence is reported separately because it deliberately supplies target gold evidence as an upper bound.

For each target, we select eight demonstrations by lexical nearest-neighbor matching using token-set overlap. The same selections are used across systems when fields are available, so the main difference is whether demonstrations contain silver evidence, gold evidence, both, or neither.

The selection procedure is applied after the example and target pools have been separated. Thus, lexical similarity can determine which eligible training examples are shown, but it cannot introduce a target example or its annotation into the prompt. Keeping the selected identifiers fixed across controlled systems also avoids attributing a performance difference to a different demonstration set. Examples whose required fields are unavailable are skipped for that system, and the policy setting uses the same full-method demonstration format because it has no expert gold-evidence annotation.

\subsection{Final Inference}

For target $(q_j,R_j)$, the final prompt asks the model to produce:
\[
(\hat{S}_j, \hat{y}_j, \hat{G}_j) =
f_{\theta}^{final}(q_j, R_j, \mathcal{D}_{ICL}).
\]
The final prompt places demonstrations before the target. Demonstrations are compact silver evidence and supervision; the target contains raw retrieval, so the prompt is asymmetric. One structured call returns silver evidence first, then the label and gold-style evidence when supported. Silver evidence is not frozen in a separate call before prediction.

\subsection{Compared Systems}

We compare the following systems:

\begin{itemize}
    \setlength{\itemsep}{0pt}
    \setlength{\parskip}{0pt}
    \setlength{\parsep}{0pt}
    \item \textbf{Vanilla RAG}: target query and retrieval only.
    \item \textbf{Gold-only ICL + RAG}: demonstrations with labels and expert evidence; the policy setting is label-only because it has no expert evidence.
    \item \textbf{Silver-only ICL + RAG}: demonstrations with silver evidence and labels.
    \item \textbf{Full method}: silver evidence, labels, and expert evidence where available.
    \item \textbf{InstructRAG-style ICL}: query, answer-conditioned rationale, and label; retrieval is retained except in budget-matched policy and HICRIC/IMR variants.
    \item \textbf{Oracle evidence}: target gold evidence supplied as an upper bound.
\end{itemize}

\subsection{Metrics}

We report label accuracy and macro-F1, plus evidence F1 and Sentence/Span F1 where annotations exist. Evidence F1 measures token/text overlap with gold or weak-gold evidence; Sentence/Span F1 measures overlap of annotated evidence units (spans for Evidence Inference/HICRIC and sentence IDs for SciFact). Policy retrieval data has no expert evidence, so evidence metrics are omitted.

All evidence scores are computed against the dataset-native reference representation rather than against the generated silver field. In particular, the silver output is an intermediate signal and is not treated as gold during evaluation. This distinction matters for interpreting the ablations: a system can generate a plausible compact summary while still failing to identify the exact sentence or span used by the annotation. We therefore report label performance and evidence overlap separately and avoid treating higher textual overlap as proof that the generated explanation caused the decision.

\section{Datasets}

\paragraph{Evidence Inference (EI) 2.0.}
EI 2.0 pairs clinical-trial prompts with comparative outcome labels and expert evidence spans. We evaluate 260 completed test examples.

\paragraph{SciFact.}
SciFact pairs claims with abstracts and sentence-level rationales, using support, contradict, and not-enough-information labels. Because test labels are hidden, we use 300 labeled validation examples.

\paragraph{Prior-authorization policy retrieval.}
The public policy dataset contains synthetic requests and CMS-derived policy chunks mapped to approve, deny, or pend. It lacks expert evidence spans, so we report labels only.

\paragraph{HICRIC/IMR.}
HICRIC/IMR combines California DMHC appeal records with HICRIC-aligned CMS passages. We remove explicit outcomes, use binary approve/deny prediction, and evaluate a 2,000-example label-balanced test subset. Evidence labels provide weak reviewer-rationale supervision.

\section{Experimental Setup}

\paragraph{Implementation.}
All systems use fixed within-dataset retrieval. Stage 1 uses one LLM call per training example to generate cached silver evidence. At final inference, Stage 2 uses one structured LLM call to generate target silver evidence and the final prediction. We use temperature 1.0, top-$p$ 1.0, one completion, no seed, and no user-specified maximum-token cap as the model parameters; ICL uses eight deterministic lexical nearest-neighbor examples based on token-set overlap and outputs follow structured JSON schemas. Table~\ref{tab:model-sensitivity-main} reports the compact backend comparison; the full comparison is in Supplementary Table~\ref{tab:model-sensitivity-full}.


\paragraph{Demonstration and target pools.}
Demonstrations and targets use disjoint prepared splits: training examples for demonstrations and the designated evaluation split for each dataset.

The retrieval stage is intentionally kept outside the main comparison. 
Each dataset uses its prepared corpus and dataset-specific retriever or passage selector, and the resulting passages are written once before any LLM call. Every system then receives the same target retrieval for a given row. This design makes the comparison about representation and prompting rather than about a change in corpus coverage. It also makes the cached silver files auditable: for any demonstration, a reviewer can inspect the original query, the retrieved passages, the silver output, and the expert fields independently.

The two LLM stages use the same source-bound task definition but different information access. Stage 1 sees a training query and its retrieval while being denied the label and gold evidence; its silver output is cached for demonstrations. Stage 2 sees the target retrieval and the selected demonstrations, including their labels and evidence fields, and jointly generates target silver evidence and the final prediction. The target label and target evidence are never inserted into the prompt. Structured parsing rejects malformed or empty evidence items and invalid labels before metrics are computed. 

\begin{figure*}[!t]
\centering
\includegraphics[width=\textwidth]{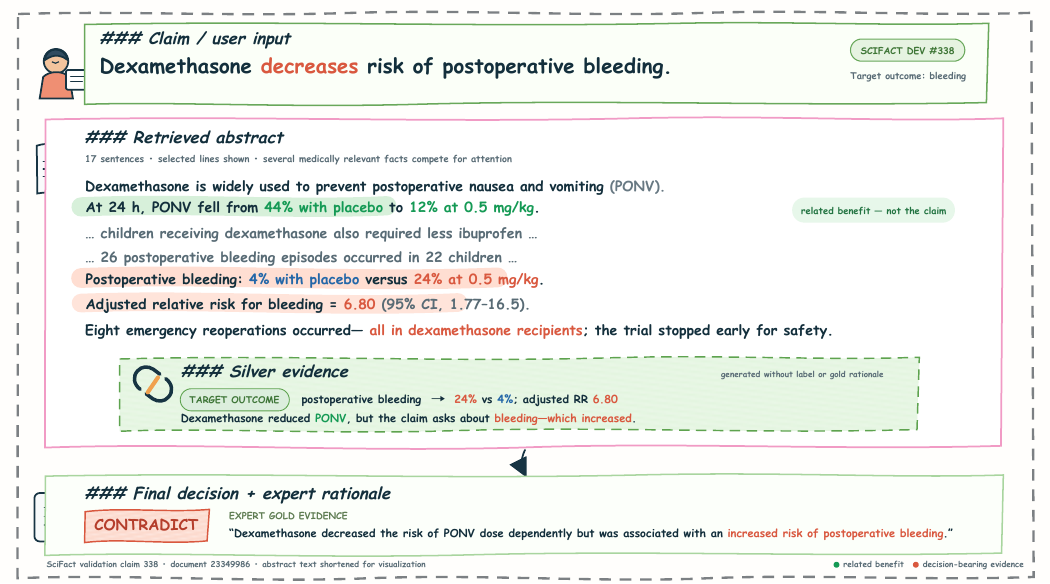}
\caption{Qualitative SciFact validation example (claim 338). The retrieved abstract contains a genuine but decision-irrelevant benefit---reduced postoperative nausea and vomiting---while the claim concerns postoperative bleeding. Silver evidence isolates the target outcome and effect size (24\% versus 4\%; adjusted relative risk 6.80) before a \textsc{Contradict} prediction aligned with the expert rationale. Text is shortened for visualization.}
\label{fig:qual-example}
\end{figure*}

\paragraph{Qualitative illustration.}
Figure~\ref{fig:qual-example} shows silver evidence filtering a misleading SciFact result: although the abstract reports reduced nausea and vomiting, the claim concerns bleeding, yielding a \textsc{Contradict} prediction aligned with the expert rationale.

\section{Results}
\label{sec:results}

GPT-5.5 has the strongest aggregate label performance in the full-method backend comparison (Table~\ref{tab:model-sensitivity-main}). Table~\ref{tab:label-results} reports its corresponding system comparison.
We treat gold-only ICL + RAG (label-only for policy) as the primary baseline because it uses the same number of demonstrations as the full method. Relative to this baseline, full-method accuracy changes by $-0.1$ points on EI 2.0 and improves by 33.0, 3.3, and 8.0 points on policy retrieval, HICRIC/IMR, and SciFact. It has higher mean accuracy than InstructRAG-style ICL on all four datasets; on policy retrieval, InstructRAG has slightly higher Macro-F1. Vanilla RAG is a lower-bound reference rather than the primary controlled comparison. We do not conduct paired significance tests, so small differences should be interpreted cautiously.

\begin{table}[!t]
\centering
\small
\setlength{\tabcolsep}{5pt}
\resizebox{\columnwidth}{!}{%
\begin{tabular}{llcc}
\toprule
Dataset & System & Accuracy & Macro-F1 \\
\midrule
\multirow{6}{*}{EI 2.0}
& Vanilla RAG & $91.2 \pm 0.3$ & $91.0 \pm 0.8$ \\
& Gold-only ICL + RAG & $\mathbf{91.5 \pm 0.2}$ & $\mathbf{91.3 \pm 0.7}$ \\
& Silver-only ICL + RAG & $91.2 \pm 0.3$ & $91.0 \pm 0.7$ \\
& Full method & $91.4 \pm 0.1$ & $90.9 \pm 0.4$ \\
& InstructRAG-style ICL & $90.8 \pm 0.3$ & $90.5 \pm 0.9$ \\
& Oracle evidence & $93.1 \pm 0.3$ & $92.8 \pm 0.1$ \\
\midrule
\multirow{6}{*}{Policy retrieval}
& Vanilla RAG & $74.0 \pm 0.4$ & $56.5 \pm 1.0$ \\
& Label-only ICL + RAG & $66.5 \pm 0.5$ & $63.3 \pm 0.3$ \\
& Silver-only ICL + RAG & -- & -- \\
& Full method & $\mathbf{99.5 \pm 0.2}$ & $99.0 \pm 0.4$ \\
& Compressed InstructRAG-ICL & $99.0 \pm 0.1$ & $\mathbf{99.2 \pm 0.2}$ \\
& Oracle evidence & $100.0 \pm 0.0$ & $100.0 \pm 0.0$ \\
\midrule
\multirow{6}{*}{HICRIC/IMR appeals}
& Vanilla RAG & $74.1 \pm 0.3$ & $70.5 \pm 1.0$ \\
& Gold-only ICL + RAG & $86.2 \pm 0.6$ & $85.3 \pm 1.0$ \\
& Silver-only ICL + RAG & $79.2 \pm 0.2$ & $75.7 \pm 1.1$ \\
& Full method & $\mathbf{89.5 \pm 1.3}$ & $\mathbf{89.4 \pm 1.1}$ \\
& Compressed InstructRAG-ICL & $84.2 \pm 0.5$ & $82.3 \pm 0.1$ \\
& Oracle evidence & $98.1 \pm 0.8$ & $98.0 \pm 0.3$ \\
\midrule
\multirow{6}{*}{SciFact}
& Vanilla RAG & $70.0 \pm 0.5$ & $69.5 \pm 0.8$ \\
& Gold-only ICL + RAG & $73.0 \pm 1.2$ & $72.7 \pm 1.2$ \\
& Silver-only ICL + RAG & $75.0 \pm 0.6$ & $74.9 \pm 0.7$ \\
& Full method & $\mathbf{81.0 \pm 0.2}$ & $\mathbf{81.0 \pm 0.5}$ \\
& InstructRAG-style ICL & $78.0 \pm 1.3$ & $77.9 \pm 0.7$ \\
& Oracle evidence & $91.3 \pm 0.8$ & $91.7 \pm 0.3$ \\
\bottomrule
\end{tabular}
}
\caption{Label prediction results for the primary GPT-5.5 experiments. Values are percentages reported as mean $\pm$ standard deviation over three runs. The InstructRAG-style rows use full demonstration-side retrieval for EI 2.0 and SciFact and a compressed, budget-matched format for policy retrieval and HICRIC/IMR. For policy retrieval, silver-only ICL + RAG is identical to the full method because no expert gold evidence is available, so duplicate values are omitted. Oracle-evidence rows are upper bounds rather than deployable systems.}
\label{tab:label-results}
\end{table}

Table~\ref{tab:evidence-results} summarizes evidence overlap. Relative to gold-only ICL, full-method Evidence F1 gains are 3.8, 8.0, and 3.2 points on EI 2.0, SciFact, and HICRIC/IMR; corresponding Sentence/Span gains are 4.5, 8.7, and 2.2. It exceeds InstructRAG on both metrics, although HICRIC/IMR absolute overlap remains low under weak reviewer-rationale supervision.

The separation between the two evidence measures is useful for interpreting these results. Textual Evidence F1 rewards overlap with the content of the reference rationale, whereas Sentence/Span F1 rewards selecting the annotated evidence unit. A model can therefore obtain reasonable textual overlap while selecting only part of the correct span, or identify a correct sentence while paraphrasing it. The full method improves both measures on each evidence-annotated dataset, but the remaining gap to oracle evidence shows that retrieval and evidence localization are still unresolved.

\begin{table}[!t]
\centering
\scriptsize
\setlength{\tabcolsep}{2pt}
\resizebox{\columnwidth}{!}{%
\begin{tabular}{llcc}
\toprule
Dataset & System & Evidence F1 & Sentence/Span F1 \\
\midrule
\multirow{6}{*}{EI 2.0}
& Vanilla RAG & $55.7 \pm 0.4$ & $54.7 \pm 0.9$ \\
& Gold-only ICL + RAG & $58.4 \pm 0.2$ & $56.4 \pm 0.1$ \\
& Silver-only ICL + RAG & $61.7 \pm 0.9$ & $60.3 \pm 1.0$ \\
& Full method & $\mathbf{62.2 \pm 1.0}$ & $\mathbf{60.9 \pm 0.4}$ \\
& InstructRAG-style ICL & $43.5 \pm 0.9$ & $46.1 \pm 0.3$ \\
& Oracle evidence & $95.8 \pm 0.5$ & $95.4 \pm 1.3$ \\
\midrule
\multirow{6}{*}{SciFact}
& Vanilla RAG & $54.6 \pm 0.6$ & $54.5 \pm 0.4$ \\
& Gold-only ICL + RAG & $56.1 \pm 0.4$ & $56.1 \pm 1.0$ \\
& Silver-only ICL + RAG & $37.8 \pm 0.1$ & $37.7 \pm 0.3$ \\
& Full method & $\mathbf{64.1 \pm 0.5}$ & $\mathbf{64.8 \pm 0.2}$ \\
& InstructRAG-style ICL & $51.9 \pm 1.2$ & $51.8 \pm 1.0$ \\
& Oracle evidence & $99.5 \pm 1.0$ & $99.5 \pm 0.3$ \\
\midrule
\multirow{6}{*}{HICRIC/IMR appeals}
& Vanilla RAG & $21.8 \pm 0.3$ & $11.1 \pm 0.8$ \\
& Gold-only ICL + RAG & $35.2 \pm 1.0$ & $22.9 \pm 0.2$ \\
& Silver-only ICL + RAG & $19.7 \pm 0.2$ & $15.4 \pm 1.1$ \\
& Full method & $\mathbf{38.4 \pm 0.1}$ & $\mathbf{25.1 \pm 1.1}$ \\
& Compressed InstructRAG-ICL & $10.1 \pm 0.3$ & $8.5 \pm 1.1$ \\
& Oracle evidence & $92.1 \pm 0.2$ & $89.7 \pm 0.9$ \\
\bottomrule
\end{tabular}
}
\caption{Evidence-overlap results for the primary GPT-5.5 experiments. Values are percentages reported as mean $\pm$ standard deviation over three runs. Evidence F1 is token/rationale F1 for EI 2.0 and HICRIC/IMR and text F1 for SciFact. Sentence/Span F1 is span F1 for EI 2.0 and HICRIC/IMR and sentence-id F1 for SciFact. HICRIC/IMR uses weak reviewer-rationale evidence rather than expert span annotation. Policy retrieval lacks gold evidence spans and is omitted.}
\label{tab:evidence-results}
\end{table}

\paragraph{Model sensitivity.}
Table~\ref{tab:model-sensitivity-main} shows dataset-dependent backend rankings: GPT-5.6-Terra and Gemini Pro 3.1 are strongest on EI 2.0 label metrics, GPT-5.5 on SciFact, policy, and HICRIC labels, Gemini Flash 3.7 on EI 2.0 evidence metrics, and Claude Opus 4.8 on HICRIC evidence metrics. No backend dominates every metric. We use GPT-5.5 for the primary tables because it provides the strongest aggregate label performance across the four datasets; the full backend comparison, including other GPT model variants, appears in Supplementary Table~\ref{tab:model-sensitivity-full}.

\begin{table*}[!t]
\centering
\small
\setlength{\tabcolsep}{4pt}
\resizebox{\textwidth}{!}{%
\begin{tabular}{llcccccc}
\toprule
Dataset & LLM backend & Accuracy & Macro-F1 & Evidence F1 & Sentence/Span F1 & Total tokens & Avg. latency (s) \\
\midrule
\multirow{8}{*}{EI 2.0} & GPT-5.5 & $91.4 \pm 0.1$ & $90.9 \pm 0.4$ & $62.2 \pm 1.0$ & $60.9 \pm 0.4$ & $2.19$M & $11.7$ \\
                                    & GPT-5.6-Sol & $90.0 \pm 0.8$ & $89.6 \pm 0.9$ & $64.7 \pm 0.2$ & $64.0 \pm 1.1$ & $1.78$M & $4.9$ \\
                                    & GPT-5.6-Terra & $\mathbf{91.9 \pm 1.0}$ & $\mathbf{91.7 \pm 1.1}$ & $66.7 \pm 0.4$ & $65.0 \pm 0.9$ & $2.09$M & $11.1$ \\
                                    & GPT-5.6-Luna & $90.0 \pm 0.4$ & $89.6 \pm 0.7$ & $62.3 \pm 1.3$ & $61.5 \pm 0.8$ & $1.80$M & $3.7$ \\
                                    & Gemini Pro 3.1 & $\mathbf{91.9 \pm 0.2}$ & $\mathbf{91.7 \pm 1.2}$ & $66.7 \pm 0.1$ & $65.0 \pm 0.3$ & $2.09$M & $11.1$ \\
                                    & Gemini Flash 3.7 & $90.4 \pm 0.2$ & $90.2 \pm 0.5$ & $\mathbf{67.3 \pm 0.8}$ & $\mathbf{66.2 \pm 0.7}$ & $2.00$M & $6.5$ \\
                                    & Claude Sonnet 5 & $90.9 \pm 0.2$ & $90.4 \pm 0.2$ & $64.1 \pm 0.3$ & $62.7 \pm 0.2$ & $2.22$M & $7.5$ \\
                                    & Claude Opus 4.8 & $90.5 \pm 0.2$ & $90.2 \pm 0.3$ & $66.8 \pm 0.1$ & $65.5 \pm 0.4$ & $2.30$M & $8.1$ \\
\midrule
\multirow{8}{*}{SciFact}            & GPT-5.5 & $\mathbf{81.0 \pm 0.2}$ & $\mathbf{81.0 \pm 0.5}$ & $\mathbf{64.1 \pm 0.5}$ & $\mathbf{64.8 \pm 0.2}$ & $1.86$M & $19.3$ \\
                                    & GPT-5.6-Sol & $77.7 \pm 0.9$ & $77.6 \pm 0.7$ & $60.3 \pm 1.0$ & $60.7 \pm 0.5$ & $1.62$M & $6.4$ \\
                                    & GPT-5.6-Terra & $71.7 \pm 0.5$ & $71.1 \pm 1.3$ & $58.6 \pm 0.6$ & $58.6 \pm 0.5$ & $1.86$M & $8.7$ \\
                                    & GPT-5.6-Luna & $74.7 \pm 1.2$ & $74.8 \pm 0.1$ & $57.9 \pm 0.3$ & $57.8 \pm 0.9$ & $1.62$M & $4.4$ \\
                                    & Gemini Pro 3.1 & $71.7 \pm 1.1$ & $71.1 \pm 0.3$ & $58.6 \pm 0.9$ & $58.6 \pm 0.5$ & $1.86$M & $8.7$ \\
                                    & Gemini Flash 3.7 & $75.3 \pm 0.6$ & $75.0 \pm 0.3$ & $62.2 \pm 0.6$ & $62.4 \pm 0.7$ & $1.79$M & $6.9$ \\
                                    & Claude Sonnet 5 & $73.8 \pm 0.7$ & $73.4 \pm 0.6$ & $57.5 \pm 0.9$ & $57.5 \pm 0.9$ & $2.00$M & $6.7$ \\
                                    & Claude Opus 4.8 & $77.6 \pm 0.0$ & $77.4 \pm 0.0$ & $61.0 \pm 0.0$ & $61.0 \pm 0.0$ & $2.08$M & $6.3$ \\
\midrule
\multirow{8}{*}{Policy retrieval}   & GPT-5.5 & $\mathbf{99.5 \pm 0.2}$ & $\mathbf{99.0 \pm 0.4}$ & -- & -- & $1.50$M & $6.3$ \\
                                    & GPT-5.6-Sol & $99.0 \pm 0.1$ & $97.9 \pm 0.3$ & -- & -- & $1.30$M & $3.8$ \\
                                    & GPT-5.6-Terra & $95.0 \pm 1.1$ & $93.4 \pm 0.2$ & -- & -- & $1.51$M & $11.1$ \\
                                    & GPT-5.6-Luna & $91.5 \pm 1.1$ & $85.7 \pm 0.5$ & -- & -- & $1.32$M & $3.3$ \\
                                    & Gemini Pro 3.1 & $95.0 \pm 1.0$ & $93.4 \pm 0.9$ & -- & -- & $1.51$M & $11.1$ \\
                                    & Gemini Flash 3.7 & $99.0 \pm 0.1$ & $\mathbf{99.0 \pm 0.4}$ & -- & -- & $1.55$M & $6.1$ \\
                                    & Claude Sonnet 5 & $95.3 \pm 0.4$ & $92.0 \pm 0.6$ & -- & -- & $3.00$M & $9.3$ \\
                                    & Claude Opus 4.8 & $98.0 \pm 0.5$ & $95.4 \pm 1.2$ & -- & -- & $3.10$M & $8.8$ \\
\midrule
\multirow{8}{*}{HICRIC/IMR appeals} & GPT-5.5 & $\mathbf{89.5 \pm 1.3}$ & $\mathbf{89.4 \pm 1.1}$ & $38.4 \pm 0.1$ & $25.1 \pm 1.1$ & $32.67$M & $16.4$ \\
                                    & GPT-5.6-Sol & $89.0 \pm 1.3$ & $89.0 \pm 0.1$ & $33.4 \pm 0.9$ & $23.2 \pm 0.3$ & $29.01$M & $5.9$ \\
                                    & GPT-5.6-Terra & $88.5 \pm 0.4$ & $88.5 \pm 0.4$ & $33.0 \pm 0.5$ & $23.5 \pm 0.9$ & $28.98$M & $5.2$ \\
                                    & GPT-5.6-Luna & $87.0 \pm 0.8$ & $87.0 \pm 0.1$ & $36.0 \pm 0.2$ & $26.4 \pm 0.6$ & $29.83$M & $5.4$ \\
                                    & Gemini Pro 3.1 & $88.0 \pm 0.9$ & $87.9 \pm 0.2$ & $32.1 \pm 0.5$ & $19.5 \pm 1.0$ & $30.91$M & $5.2$ \\
                                    & Gemini Flash 3.7 & $87.6 \pm 0.4$ & $87.5 \pm 0.5$ & $38.8 \pm 0.9$ & $30.8 \pm 0.6$ & $34.94$M & $7.3$ \\
                                    & Claude Sonnet 5 & $85.6 \pm 0.0$ & $85.5 \pm 0.0$ & $37.3 \pm 0.0$ & $28.7 \pm 0.0$ & $61.45$M & $11.9$ \\
                                    & Claude Opus 4.8 & $87.6 \pm 0.0$ & $87.6 \pm 0.0$ & $\mathbf{39.7 \pm 0.0}$ & $\mathbf{32.5 \pm 0.0}$ & $57.79$M & $10.5$ \\
\bottomrule
\end{tabular}}
\caption{Full-method backend sensitivity for the main-paper comparison. The complete backend table appears in Supplementary Table~\ref{tab:model-sensitivity-full}. Accuracy, Macro-F1, Evidence F1, and Sentence/Span F1 are reported as mean $\pm$ standard deviation. Total-token and latency columns report means across runs without dispersion terms. Bold identifies the best available result for each dataset and metric; ties are retained.}
\label{tab:model-sensitivity-main}
\end{table*}

Table~\ref{tab:sota-comparison} gives context rather than a leaderboard: splits, decision protocols, and label spaces differ, especially for SciFact, policy retrieval, and adapted HICRIC/IMR. We therefore report label-level comparisons only.

\begin{table}[!t]
\centering
\scriptsize
\setlength{\tabcolsep}{2pt}
\resizebox{\columnwidth}{!}{%
\begin{tabular}{llcc}
\toprule
Dataset & Published comparator / setting & Published & Full method \\
\midrule
EI 2.0 & Biomed RoBERTa macro-F1 \cite{deyoung2020evidence} & 78.0 & \textbf{90.9} \\
SciFact & MultiVerS label F1 \cite{wadden2022multivers} & 72.5 & \textbf{81.0}$^\dagger$ \\
Policy & Adaptive DPO accuracy \cite{sharifullin2026offline} & 92.0 & \textbf{99.5} \\
Policy & Fixed-$K$ accuracy \cite{sharifullin2026offline} & 62.0 & \textbf{99.5} \\
HICRIC & ClinicalBERT accuracy \cite{gartner2024hicric} & 74.0 & \textbf{89.5}$^\ddagger$ \\
HICRIC & DistilBERT macro-F1 \cite{gartner2024hicric} & 72.7 & \textbf{89.4}$^\ddagger$ \\
\bottomrule
\end{tabular}
}
\caption{Contextual label-level comparison with published task-specific systems. Values are percentages; the Full method entry uses the metric named by each comparator row. Rows are not strictly comparable because retrieval setting, split, label space, and evaluation protocol differ. $^\dagger$SciFact is evaluated on the labeled validation split because official test labels are not public. $^\ddagger$HICRIC/IMR uses our adapted binary approve/deny appeal-policy setting rather than the original three-class benchmark.}
\label{tab:sota-comparison}
\end{table}

\section{Ablation Study}

\paragraph{Vanilla RAG as a lower bound.}
The full method improves over zero-demonstration vanilla RAG in all four settings, but this combines demonstration, silver, and expert-supervision effects. Gold-only ICL + RAG (label-only for policy), with eight demonstrations, is the primary controlled baseline.

\paragraph{Silver evidence vs. gold-/label-only ICL.}
Gold-only ICL controls demonstration count and expert supervision on the annotated datasets; policy uses a label-only counterpart. Relative accuracy changes are $-0.1$, +33.0, +3.3, and +8.0 points on EI 2.0, policy, HICRIC/IMR, and SciFact; Evidence F1 gains are 3.8, 8.0, and 3.2 points on the annotated datasets. The policy and SciFact results suggest that the generated representation is especially useful when retrieval must be filtered or transformed, while the small EI 2.0 difference shows that compact clinical evidence already supports strong label prediction.

\paragraph{Full method vs. silver-only ICL+RAG.}
Adding gold evidence to silver demonstrations changes accuracy by 0.2, 6.0, and 10.3 points on EI 2.0, SciFact, and HICRIC/IMR. HICRIC/IMR rises from 79.2\% accuracy and 19.7 Evidence F1 to 89.5\% and 38.4, while SciFact rises from 37.8 to 64.1 Evidence F1. These results indicate that generated summaries alone do not uniformly provide the decision boundary; expert evidence remains important for the final evidence-oriented task.

\paragraph{Comparison with InstructRAG-style ICL.}
The full method exceeds InstructRAG by 0.6, 3.0, and 5.3 accuracy points on EI 2.0, SciFact, and HICRIC/IMR and produces higher evidence overlap on all three evidence-annotated datasets. On policy, InstructRAG has slightly higher Macro-F1 (99.2\% vs. 99.0\%) but lower accuracy (99.0\% vs. 99.5\%). Answer-conditioned rationales may therefore help label prediction, whereas the silver format is competitive when evidence localization is evaluated. The comparison is informative but not a pure label-conditioning ablation because demonstration retrieval differs.

\paragraph{Evidence-quality metrics.}
The evidence metrics clarify where the benefit lies. EI 2.0 rises from 58.4 to 62.2 Evidence F1 and 56.4 to 60.9 Span F1 despite stable labels. SciFact rises from 56.1 to 64.1 Evidence F1 and 56.1 to 64.8 Sentence F1, while silver-only is 37.8. HICRIC/IMR rises from 35.2 to 38.4 and 22.9 to 25.1, but absolute overlap remains limited. These results support combining generated and expert evidence, but do not establish causal mediation.

\paragraph{Oracle gap.}
Oracle evidence remains stronger, especially on SciFact (91.3\% vs. 81.0\%) and HICRIC/IMR (98.1\% vs. 89.5\%), showing that retrieval and evidence selection remain bottlenecks. 

\paragraph{Resource use.} GPT-5.5 full-method runs use 2.19M, 1.86M, 1.50M, and 32.67M tokens on EI 2.0, SciFact, policy retrieval, and HICRIC/IMR, respectively. These totals reflect differences in dataset size and context length and are not directly comparable across datasets. Compact demonstrations may reduce inference-prompt length, while offline silver-evidence generation adds preprocessing cost.

\section{Discussion}

The experiments support a narrower claim than comparison with vanilla RAG alone would suggest. Silver evidence is a useful asymmetric representation that replaces demonstration-side retrieval with generated evidence paired with expert supervision. However, because demonstrations do not retain paired raw retrieval and silver evidence, and test-time silver evidence and labels are generated in one call, the experiments do not establish a causal retrieval-to-evidence-to-decision mechanism. Compression, output structure, and intermediate supervision remain partly confounded.

The largest accuracy gains occur on policy and SciFact, though likely for different reasons: policy may benefit from filtering long, noisy criteria, whereas SciFact may benefit from isolating evidence relevant to fine-grained entailment; gains are smaller on HICRIC/IMR and minimal on EI 2.0.
InstructRAG has slightly higher policy Macro-F1, while the full method has higher accuracy and stronger evidence results on annotated datasets. In coverage and prior-authorization settings, evidence remains important for audit and review.

The EI 2.0 and SciFact results reveal different boundary conditions. Compact EI 2.0 passages leave little room for label gains despite improved evidence quality. SciFact shows substantial gains, but generated summaries may blur fine-grained distinctions among support, contradiction, and insufficient information. Silver evidence is most reliable when it removes distraction without erasing decision-critical distinctions.

The contrast across datasets helps distinguish a representation benefit from a universal reasoning improvement. Policy retrieval exposes long, heterogeneous criteria in which a compact decision-oriented representation can remove distraction before the model applies the label boundary. HICRIC/IMR has a similar document-grounding requirement, but its adapted binary task and weak reviewer rationales make evidence matching harder. EI 2.0 provides expert spans but often presents a compact intervention--outcome comparison, so the model has less need for compression. SciFact presents the opposite challenge: the relevant abstract sentences are short, but the label depends on subtle entailment, directionality, population, and outcome distinctions. In that setting, an incorrect or over-general silver summary can erase precisely the distinction needed to separate support, contradiction, and insufficient information.

\section{Conclusion}

We introduced an asymmetric silver-evidence demonstration format for retrieval-augmented ICL: demonstrations replace raw retrieval with label-blind summaries paired with expert supervision, whereas test inputs retain raw retrieved passages. At test time, one structured call generates silver evidence from the raw retrieval before predicting the label and decision evidence. Our full method improves labels on three of four datasets and evidence overlap on all annotated datasets, while achieving higher label accuracy than InstructRAG on all four.

\clearpage

\appendix
\section{Supplementary Material}

\subsection{Prompting Details}

The scripts use dataset-specific field names and labels but the following common structure. Deployable systems receive retrieval at test time, never target labels or gold evidence, and return parsed structured JSON.

\paragraph{Shared system instruction.}
Both LLM stages use a task-specific evidence-grounding instruction with the following behavior:

\begin{quote}
\small
You are an evidence-focused decision-support assistant. Use only the provided target query and retrieved passages. Do not rely on outside knowledge. Extract evidence that is directly present in, or tightly supported by, the retrieved text. Keep evidence concise, preserve important titles, section headers, sentence identifiers, or policy headings when they are provided, and return only valid JSON following the requested schema.
\end{quote}

This source-bound instruction separates decision-relevant evidence from distracting retrieval and discourages unsupported text.

\paragraph{Stage 1: silver-evidence generation.}
For each demonstration-pool example, Stage 1 receives only query and retrieval:
\[
\text{input}_{silver}=(q_i,R_i).
\]
The gold label $y_i$ and evidence $G_i$ are excluded. The prompt is:

\begin{quote}
\small
\textbf{Task.} Generate concise silver evidence from the retrieved passages. Do not predict or infer the label or expert evidence. Select only information useful for the downstream decision and preserve available titles, headers, categories, sentence IDs, or policy sections.

\textbf{Target query.} \emph{\{dataset-specific claim, clinical question, policy request, or appeal background\}}

\textbf{Retrieved passages.} \emph{\{passage id, source, title/header if available, text\}}

\textbf{Returned JSON.} \emph{\{silver\_evidence, explanation\}}
\end{quote}

When metadata are available, evidence is header-keyed:
\begin{quote}
\small
\texttt{\{"heading": "...", "evidence": "...", "source\_ids": ["..."]\}}
\end{quote}
Without reliable headings, the field is omitted or uses a short source label. Silver evidence is generated once and cached for later runs.

\paragraph{Stage 2: final inference with silver-evidence demonstrations.}
Stage 2 receives ICL demonstrations and target retrieval. A full demonstration is:
\[
(q_i,S_i,y_i,G_i),
\]
where $S_i$ is cached and $G_i$ is included when available. Demonstrations replace raw retrieval; the target retains raw retrieval:

\begin{quote}
\small
\textbf{Demonstration.} Query; silver evidence; label; gold evidence if available.

\textbf{Target example.} Target query; retrieved passages.

\textbf{Instruction.} For the target, first generate silver evidence, then predict the label and, where applicable, gold-style decision evidence. Return valid JSON.
\end{quote}

The output schema is:
\begin{quote}
\small
\texttt{\{silver\_evidence, label, gold\_evidence, explanation\}}
\end{quote}
This is one structured call; test-time silver evidence is not frozen in a separate call.

\subsection{Baseline Prompt Formats}

\paragraph{Vanilla RAG.}
The target query and retrieval are provided without demonstrations:
\[
(q_j,R_j)\rightarrow(\hat{y}_j,\hat{G}_j).
\]

\paragraph{Gold-only/label-only ICL + RAG.}
Demonstrations contain query, expert label, and gold evidence but no silver field; the target retains retrieval. Policy uses the label-only form because expert evidence is unavailable.

\paragraph{Silver-only ICL + RAG.}
Demonstrations contain query, cached silver evidence, and label but omit gold evidence.

\paragraph{Full method.}
Demonstrations contain query, cached silver evidence, expert label, and gold evidence when available; the target contains retrieval.

\paragraph{Oracle evidence.}
The target receives gold evidence directly as a non-deployable upper bound.

\paragraph{InstructRAG-style ICL.}
Stage 1 generates a label-conditioned rationale. Demonstrations include query, retrieval, rationale, and label; unlike our method, retrieval is retained. We use a compressed, budget-matched variant for policy and HICRIC/IMR and retain retrieval for EI 2.0 and SciFact. Results appear in Tables~\ref{tab:label-results}--\ref{tab:evidence-results}.

\subsection{Dataset-Specific Prompt Fields}

\paragraph{EI 2.0.}
The query contains intervention, comparator, outcome, and available article metadata; retrieval supplies article snippets, and evidence is scored against expert spans.

\paragraph{SciFact.}
The query is a scientific claim; retrieval supplies titled abstract sentences with document and sentence IDs. Labels are \textsc{Support}, \textsc{Contradict}, and \textsc{Not Enough Info}; evidence uses rationale IDs and text overlap.

\paragraph{Prior-authorization policy retrieval.}
The query is a prior-authorization request and decision question; retrieval supplies policy criteria. Labels are approve, deny, or pend. No expert evidence spans are available.

\paragraph{HICRIC/IMR appeals.}
The query uses IMR treatment, diagnosis, background, and available rationale; retrieval supplies HICRIC-aligned CMS passages. Labels are approve/deny, and evidence uses weak rationale text.

\subsection{Backend Identifiers}

Table~\ref{tab:backend-identifiers} reports the provider-facing model and deployment identifiers used in the experiments. Identifiers and API access were last verified on October 12, 2026.

\begin{table*}[t]
\centering
\small
\setlength{\tabcolsep}{4pt}
\resizebox{\textwidth}{!}{%
\begin{tabular}{llll}
\toprule
Display name & Provider & API model identifier & Configured deployment identifier \\
\midrule
GPT-4.1 & OpenAI & \texttt{gpt-4.1} & \texttt{gpt-4.1\_2025-04-14\_no-guardrails} \\
GPT-5.2 & OpenAI & \texttt{gpt-5.2} & \texttt{gpt-5.2\_2025-12-11} \\
GPT-5.5 & OpenAI & \texttt{gpt-5.5} & \texttt{gpt-5.5\_2026-04-24} \\
GPT-5.6-Sol & OpenAI & \texttt{gpt-5.6-sol} & \texttt{gpt-5.6-sol\_2026-07-09} \\
GPT-5.6-Terra & OpenAI & \texttt{gpt-5.6-terra} & \texttt{gpt-5.6-terra\_2026-07-09} \\
GPT-5.6-Luna & OpenAI & \texttt{gpt-5.6-luna} & \texttt{gpt-5.6-luna\_2026-07-09} \\
Gemini Pro 3.1 & Google & \texttt{gemini-3.1-pro} & \texttt{gemini-3.1-pro} \\
Gemini Flash 3.7 & Google & \texttt{gemini-3.7-flash} & \texttt{gemini-3.7-flash} \\
Claude Sonnet 5 & Anthropic & \texttt{us.anthropic.claude-sonnet-5} & \texttt{us.anthropic.claude-sonnet-5} \\
Claude Opus 4.8 & Anthropic & \texttt{us.anthropic.claude-opus-4-8} & \texttt{us.anthropic.claude-opus-4-8} \\
\bottomrule
\end{tabular}}
\caption{Provider-facing model and configured deployment identifiers used in the backend comparison.}
\label{tab:backend-identifiers}
\end{table*}

\paragraph{Model-sensitivity results.}
The complete backend comparison appears below in Supplementary Table~\ref{tab:model-sensitivity-full}.

\begin{table*}[t]
\centering
\small
\setlength{\tabcolsep}{4pt}
\resizebox{\textwidth}{!}{%
\begin{tabular}{llcccccc}
\toprule
Dataset & LLM backend & Accuracy & Macro-F1 & Evidence F1 & Sentence/Span F1 & Total tokens & Avg. latency (s) \\
\midrule
\multirow{10}{*}{EI 2.0} & GPT-4.1 & $88.8 \pm 1.2$ & $88.3 \pm 1.1$ & $62.2 \pm 1.1$ & $61.1 \pm 0.3$ & $2.01$M & $3.6$ \\
                                    & GPT-5.2 & $88.5 \pm 0.8$ & $87.8 \pm 1.2$ & $60.6 \pm 0.3$ & $58.0 \pm 0.6$ & $2.07$M & $5.0$ \\
                                    & GPT-5.5 & $91.4 \pm 0.1$ & $90.9 \pm 0.4$ & $62.2 \pm 1.0$ & $60.9 \pm 0.4$ & $2.19$M & $11.7$ \\
                                    & GPT-5.6-Sol & $90.0 \pm 0.8$ & $89.6 \pm 0.9$ & $64.7 \pm 0.2$ & $64.0 \pm 1.1$ & $1.78$M & $4.9$ \\
                                    & GPT-5.6-Terra & $\mathbf{91.9 \pm 1.0}$ & $\mathbf{91.7 \pm 1.1}$ & $66.7 \pm 0.4$ & $65.0 \pm 0.9$ & $2.09$M & $11.1$ \\
                                    & GPT-5.6-Luna & $90.0 \pm 0.4$ & $89.6 \pm 0.7$ & $62.3 \pm 1.3$ & $61.5 \pm 0.8$ & $1.80$M & $3.7$ \\
                                    & Gemini Pro 3.1 & $\mathbf{91.9 \pm 0.2}$ & $\mathbf{91.7 \pm 1.2}$ & $66.7 \pm 0.1$ & $65.0 \pm 0.3$ & $2.09$M & $11.1$ \\
                                    & Gemini Flash 3.7 & $90.4 \pm 0.2$ & $90.2 \pm 0.5$ & $\mathbf{67.3 \pm 0.8}$ & $\mathbf{66.2 \pm 0.7}$ & $2.00$M & $6.5$ \\
                                    & Claude Sonnet 5 & $90.9 \pm 0.2$ & $90.4 \pm 0.2$ & $64.1 \pm 0.3$ & $62.7 \pm 0.2$ & $2.22$M & $7.5$ \\
                                    & Claude Opus 4.8 & $90.5 \pm 0.2$ & $90.2 \pm 0.3$ & $66.8 \pm 0.1$ & $65.5 \pm 0.4$ & $2.30$M & $8.1$ \\
\midrule
\multirow{10}{*}{SciFact}            & GPT-4.1 & $75.9 \pm 0.6$ & $75.5 \pm 0.5$ & $58.1 \pm 0.3$ & $58.1 \pm 0.3$ & $1.79$M & $4.1$ \\
                                    & GPT-5.2 & $70.7 \pm 0.2$ & $70.9 \pm 0.2$ & $57.3 \pm 0.3$ & $57.3 \pm 0.5$ & $1.87$M & $5.1$ \\
                                    & GPT-5.5 & $\mathbf{81.0 \pm 0.2}$ & $\mathbf{81.0 \pm 0.5}$ & $\mathbf{64.1 \pm 0.5}$ & $\mathbf{64.8 \pm 0.2}$ & $1.86$M & $19.3$ \\
                                    & GPT-5.6-Sol & $77.7 \pm 0.9$ & $77.6 \pm 0.7$ & $60.3 \pm 1.0$ & $60.7 \pm 0.5$ & $1.62$M & $6.4$ \\
                                    & GPT-5.6-Terra & $71.7 \pm 0.5$ & $71.1 \pm 1.3$ & $58.6 \pm 0.6$ & $58.6 \pm 0.5$ & $1.86$M & $8.7$ \\
                                    & GPT-5.6-Luna & $74.7 \pm 1.2$ & $74.8 \pm 0.1$ & $57.9 \pm 0.3$ & $57.8 \pm 0.9$ & $1.62$M & $4.4$ \\
                                    & Gemini Pro 3.1 & $71.7 \pm 1.1$ & $71.1 \pm 0.3$ & $58.6 \pm 0.9$ & $58.6 \pm 0.5$ & $1.86$M & $8.7$ \\
                                    & Gemini Flash 3.7 & $75.3 \pm 0.6$ & $75.0 \pm 0.3$ & $62.2 \pm 0.6$ & $62.4 \pm 0.7$ & $1.79$M & $6.9$ \\
                                    & Claude Sonnet 5 & $73.8 \pm 0.7$ & $73.4 \pm 0.6$ & $57.5 \pm 0.9$ & $57.5 \pm 0.9$ & $2.00$M & $6.7$ \\
                                    & Claude Opus 4.8 & $77.6 \pm 0.0$ & $77.4 \pm 0.0$ & $61.0 \pm 0.0$ & $61.0 \pm 0.0$ & $2.08$M & $6.3$ \\
\midrule
\multirow{10}{*}{Policy retrieval}   & GPT-4.1 & $93.2 \pm 0.1$ & $90.2 \pm 1.1$ & -- & -- & $1.43$M & $2.8$ \\
                                    & GPT-5.2 & $88.5 \pm 1.0$ & $83.7 \pm 0.4$ & -- & -- & $1.44$M & $4.3$ \\
                                    & GPT-5.5 & $\mathbf{99.5 \pm 0.2}$ & $\mathbf{99.0 \pm 0.4}$ & -- & -- & $1.50$M & $6.3$ \\
                                    & GPT-5.6-Sol & $99.0 \pm 0.1$ & $97.9 \pm 0.3$ & -- & -- & $1.30$M & $3.8$ \\
                                    & GPT-5.6-Terra & $95.0 \pm 1.1$ & $93.4 \pm 0.2$ & -- & -- & $1.51$M & $11.1$ \\
                                    & GPT-5.6-Luna & $91.5 \pm 1.1$ & $85.7 \pm 0.5$ & -- & -- & $1.32$M & $3.3$ \\
                                    & Gemini Pro 3.1 & $95.0 \pm 1.0$ & $93.4 \pm 0.9$ & -- & -- & $1.51$M & $11.1$ \\
                                    & Gemini Flash 3.7 & $99.0 \pm 0.1$ & $\mathbf{99.0 \pm 0.4}$ & -- & -- & $1.55$M & $6.1$ \\
                                    & Claude Sonnet 5 & $95.3 \pm 0.4$ & $92.0 \pm 0.6$ & -- & -- & $3.00$M & $9.3$ \\
                                    & Claude Opus 4.8 & $98.0 \pm 0.5$ & $95.4 \pm 1.2$ & -- & -- & $3.10$M & $8.8$ \\
\midrule
\multirow{10}{*}{HICRIC/IMR appeals} & GPT-4.1 & $79.6 \pm 0.7$ & $79.1 \pm 0.3$ & $31.4 \pm 1.2$ & $19.1 \pm 1.2$ & $32.20$M & $3.8$ \\
                                    & GPT-5.2 & $83.4 \pm 0.9$ & $83.2 \pm 0.4$ & $32.3 \pm 0.9$ & $21.5 \pm 0.2$ & $31.99$M & $6.0$ \\
                                    & GPT-5.5 & $\mathbf{89.5 \pm 1.3}$ & $\mathbf{89.4 \pm 1.1}$ & $38.4 \pm 0.1$ & $25.1 \pm 1.1$ & $32.67$M & $16.4$ \\
                                    & GPT-5.6-Sol & $89.0 \pm 1.3$ & $89.0 \pm 0.1$ & $33.4 \pm 0.9$ & $23.2 \pm 0.3$ & $29.01$M & $5.9$ \\
                                    & GPT-5.6-Terra & $88.5 \pm 0.4$ & $88.5 \pm 0.4$ & $33.0 \pm 0.5$ & $23.5 \pm 0.9$ & $28.98$M & $5.2$ \\
                                    & GPT-5.6-Luna & $87.0 \pm 0.8$ & $87.0 \pm 0.1$ & $36.0 \pm 0.2$ & $26.4 \pm 0.6$ & $29.83$M & $5.4$ \\
                                    & Gemini Pro 3.1 & $88.0 \pm 0.9$ & $87.9 \pm 0.2$ & $32.1 \pm 0.5$ & $19.5 \pm 1.0$ & $30.91$M & $5.2$ \\
                                    & Gemini Flash 3.7 & $87.6 \pm 0.4$ & $87.5 \pm 0.5$ & $38.8 \pm 0.9$ & $30.8 \pm 0.6$ & $34.94$M & $7.3$ \\
                                    & Claude Sonnet 5 & $85.6 \pm 0.0$ & $85.5 \pm 0.0$ & $37.3 \pm 0.0$ & $28.7 \pm 0.0$ & $61.45$M & $11.9$ \\
                                    & Claude Opus 4.8 & $87.6 \pm 0.0$ & $87.6 \pm 0.0$ & $\mathbf{39.7 \pm 0.0}$ & $\mathbf{32.5 \pm 0.0}$ & $57.79$M & $10.5$ \\
\bottomrule
\end{tabular}}
\caption{Complete full-method backend sensitivity. Accuracy, Macro-F1, Evidence F1, and Sentence/Span F1 are reported as mean $\pm$ standard deviation when repeated runs are available; single-run summaries use $\pm 0.0$. Total-token and latency columns report means across runs without dispersion terms. Bold identifies the best available result for each dataset and metric; ties are retained.}
\label{tab:model-sensitivity-full}
\end{table*}

\subsection{Other Experimental Details}
All systems use the same within-dataset retrieval. Retrieval, silver files, ICL examples, prompts, responses, usage, latency, and predictions are saved.

Eight examples are selected deterministically by lexical nearest-neighbor token overlap; selection is not manually curated or explicitly class-balanced.
GPT-5.5 is used for both stages in the primary comparison. Defaults are temperature 1.0, top-$p$ 1.0, one completion, no seed, and no user-specified maximum-token cap; sensitivity runs reuse the same splits, retrieval, prompts, and evaluators.

Repeated summaries run final inference three times with splits, retrieval, and cached silver evidence fixed and report mean/standard deviation for accuracy, macro-F1, Evidence F1, and Sentence/Span F1. The primary GPT-5.5 tables report the same three-run summaries.

\end{document}